\documentclass[conference]{IEEEtran}
\IEEEoverridecommandlockouts
\usepackage{amsmath,amssymb}
\usepackage{booktabs}
\usepackage{graphicx}
\usepackage{multirow}
\usepackage{xcolor}
\usepackage{url}
\graphicspath{{figures/}}
\title{SafeRoute-MAS: Coverage-Aware and Risk-Calibrated Message Routing for Multi-Agent Scientific Reasoning}

\author{%
\begin{tabular}{ccc}
\begin{tabular}{c}
\textbf{Zhiming Xue\textsuperscript{*}} \\
\textit{College of Engineering} \\
\textit{Northeastern University} \\
Boston, USA \\
xue.zh@northeastern.edu
\end{tabular}
&
\begin{tabular}{c}
\textbf{Jianan Liu} \\
\textit{Independent Researcher} \\
Austin, USA \\
jiananliu2408@gmail.com
\end{tabular}
&
\begin{tabular}{c}
\textbf{Sichen Zhao} \\
\textit{College of Engineering} \\
\textit{Northeastern University} \\
Boston, USA \\
zhao.siche@northeastern.edu
\end{tabular}
\\[1.5ex]
\begin{tabular}{c}
\textbf{Weiran Yan} \\
\textit{Independent Researcher} \\
Santa Clara, USA \\
yanwr2016@gmail.com
\end{tabular}
&
\begin{tabular}{c}
\textbf{Aichen Xie} \\
\textit{Independent Researcher} \\
San Jose, USA \\
xieaichen789@gmail.com
\end{tabular}
&
\begin{tabular}{c}
\textbf{Jing Yang} \\
\textit{Olin Business School} \\
\textit{Washington University in St. Louis} \\
St. Louis, USA \\
jing.y@wustl.edu
\end{tabular}
\end{tabular}
}

\begin{document}
\maketitle

\begin{abstract}
We introduce SafeRoute-MAS, a coverage-aware, risk-calibrated procedure that keeps all specialist messages or removes one using only pre-integration information.  On a sealed cache of 5,328 committees, its source-compatible operating point gains $+1.80$ percentage points (95\% CI $[+1.13,+2.52]$) while removing 12.76\% of message tokens.  We also run 4,580 controlled integrations over a stratified 458-committee sample on a public gpt-oss-120b endpoint.  The sample estimate is $-0.09$ points; design weighting to the original workload gives $+0.94$ points (95\% CI $[-1.31,+3.50]$) and 18.72\% acted-input reduction.  A weighted token-matched comparison likewise does not establish a selector advantage ($+1.68$ points, CI $[-1.27,+5.10]$).  Only 13 of 88 cached non-neutral labels (14.8\%) retain direction by repeated-call majority.  Expected-corruption UCL is 4.60\% after weighting, but only 2 of 11 prespecified marginals pass the support/risk rule.  We therefore propose a numerical qualification with Keep-All fallback, applied here retrospectively rather than validated prospectively.
\end{abstract}


\section{Introduction}
Large language model (LLM) systems increasingly solve difficult tasks through committees of specialized agents.  Debate and round-table consensus can expose an integrator to complementary evidence \cite{du2024multiagent,chen2024reconcile}.  More communication is not uniformly beneficial, however.  A plausible but misleading message may turn a correct committee answer into an error, while redundant messages consume scarce context.  Communication pruning can reduce cost \cite{zhang2025agentprune}, but a deployed router cannot use the gold correctness of a proposal and should not sacrifice helpful minority evidence merely because it disagrees with the majority.

We ask whether a router can preserve correctness while deciding which message to ignore.  SciAgentTrace \cite{yang2026sciagenttrace} provides cached full-committee and leave-one-out (LOO) outcomes from the DHD controlled replay protocol \cite{yang2026wrongbutuseful}, allowing exact evaluation of \textsc{Keep-All} and \textsc{Prune-One} within the cache.  We never add LOO effects to simulate multiple deletions.

\begin{figure*}[t]
    \centering
    \includegraphics[width=0.68\textwidth]{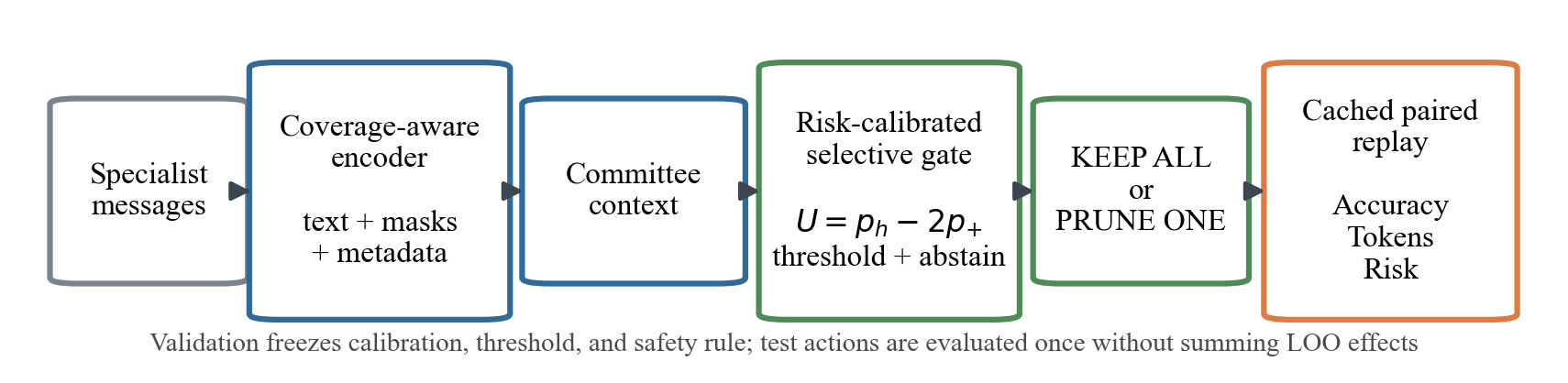}
    \caption{SafeRoute-MAS and its action-level evaluation.  Visible message and committee features feed a calibrated selective gate, which chooses \textsc{Keep-All} or one \textsc{Prune-One} action; the corresponding replay measures accuracy, token cost, and risk.  Ablations do not establish committee context or the diversity rule as independent sources of gain.}
    \label{fig:architecture}
\end{figure*}

We propose \textbf{SafeRoute-MAS}, shown in Fig.~\ref{fig:architecture}: a target-calibrated three-class message-risk model coupled to a selective gate that acts only above a validation-frozen threshold.  Our contributions are:
\begin{itemize}
    \item a leakage-safe, coverage-aware keep/prune formulation over real counterfactual actions, with explicit calibration and abstention;
    \item a numerical target-backend qualification procedure combining input validation, repetition, matched generic pruning, accuracy/token/risk gates, and Keep-All fallback; and
    \item a transparent retrospective diagnostic that finds a positive source-cache operating point but no conclusive endpoint gain or selector advantage, thereby identifying when the system should abstain.
\end{itemize}

\section{Related Work}
Multi-agent debate and consensus use interacting viewpoints to improve factuality and reasoning \cite{du2024multiagent,chen2024reconcile}.  MasRouter chooses collaboration modes, roles, and models at the system level \cite{yue2025masrouter}; SafeRoute instead selects a realized message within a fixed committee.

Communication methods prune topology \cite{zhang2025agentprune}, train efficient dialogues \cite{chen2025optima}, or prune runtime messages \cite{you2026messagesieve}; random pruning may already preserve accuracy \cite{zhang2026rethinking}.  SafeRoute instead learns one message-level action under empirical risk control, without gold correctness and with its corresponding cached replay.  AgentPrune and MessageSieve require actions outside the released cache and online system re-execution.  Calibration \cite{guo2017calibration} and selective classification \cite{geifman2017selective} motivate abstention; our UCL is not a future-distribution guarantee.

\section{Problem Formulation}
For problem $q$, committee $C_q=\{m_1,\ldots,m_K\}$ produces a cached full outcome $y_q^{\rm all}\in\{0,1\}$.  Removing message $i$ yields cached outcome $y_{q,-i}$.  Following the DHD trajectory-value intervention \cite{yang2026wrongbutuseful}, the released signed effect is
\begin{equation}
e_{qi}=y_q^{\rm all}-y_{q,-i}\in\{-1,0,+1\},
\end{equation}
where $-1$ is harmful (removal recovers the answer), $0$ is neutral, and $+1$ is helpful (removal corrupts the answer).  These outcomes are labels and evaluation fields only.

The router estimates $p_h(i)$, $p_n(i)$, and $p_+(i)$ from information visible before integration.  Candidate removal utility is
\begin{equation}
U_i=p_h(i)-\lambda p_+(i)+\gamma s_i,
\label{eq:utility}
\end{equation}
where $s_i$ is visible-character context share, $\lambda=2$, and the primary accuracy operating point fixes $\gamma=0$.  Character share requires no tokenizer at decision time; all reported savings use official tokens.  A context-weighted sensitivity uses $\gamma=0.25$.  For $i^*=\arg\max_i U_i$, the policy removes $i^*$ only if its score clears threshold $\tau$ and the frozen $K$/diversity abstention rules; otherwise it keeps all messages.  The central risk is $\Pr(e_{qi}=+1\mid\text{action})$, measured directly on selected actions.

\section{SafeRoute-MAS Design}
The full implementation retains text, committee context, and a diversity rule as frozen engineering choices.  Section~\ref{sec:ablation} shows that metadata alone is the smallest empirically sufficient representation; only calibration and observation masks have favorable point-estimate ablations, and neither is established as independently necessary.
\subsection{Coverage-Aware Message Encoding}
The encoder concatenates role name/description, reasoning lens, proposed answer, reasoning direction, key idea, critical assumption, what-to-check, and self-reported confidence.  TF--IDF word 1--2 grams represent text.  Six structured fields have separate observation bits, so an absent field is not treated as semantically empty text.  Numeric features include parse status, confidence parsed from visible language, character/word counts, and field completeness.  No reference answer, proposal correctness, judge rationale, integrator output, or effect label enters the model.

\subsection{Committee Context}
Per-message context features include normalized-answer agreement, unique/majority status, reasoning-lens frequency, relative confidence and length, context share, and committee completeness/entropy summaries.  Normalization is deterministic over raw proposal text.  Benchmark identity is included in the primary model, with a no-benchmark ablation and a LOBO evaluation.

\subsection{Calibrated Selective Gate}
A class-balanced averaged logistic SGD model predicts the three effect classes.  Message rows receive weight $1/K$ so each committee has equal training weight.  A problem-disjoint validation subset fits sigmoid calibration; the remaining validation problems select $(\tau,m)$ over a frozen threshold and diversity-margin grid.  Validation requires at least 5\% action coverage and at most 3\% observed helpful pruning.  The untouched test criterion is a 95\% problem-cluster bootstrap upper confidence limit (UCL) no greater than 5\%, with at least 100 actions over 50 acted-on problems.

The diversity rule always abstains at $K=2$.  If the selected message is the only member of its normalized-answer cluster or reasoning-lens cluster, it must clear $\tau+m$.  As Sec.~\ref{sec:ablation} shows, this conservative rule did not improve test accuracy and is retained only because it was frozen in the primary policy before test access.

Implementation uses scikit-learn \cite{pedregosa2011sklearn}: 25,000 TF--IDF 1--2 grams (\texttt{min\_df}=3, \texttt{max\_df}=.995), one-hot categories, and averaged class-balanced logistic SGD (L2 $\alpha=3\times10^{-5}$; 2,000 iterations).  Validation is split 1:2 for sigmoid calibration and gate development.  Searching 201 utility quantiles and $m\in\{0,.05,.10,.15\}$ selects $\tau=-.04558,m=.05$; $\tau$ is a ranking cutoff, not a zero-utility threshold.  Seed 20261031 is fixed.  For the cached-primary analysis, clustered CIs use 5,000 accuracy and 10,000 risk/token resamples; independent-endpoint resampling is specified in Table~\ref{tab:confirmation}.

\section{Experimental Setup}
\subsection{Data and Leakage Controls}
We freeze access-controlled SciAgentTrace v2.1.0 \cite{yang2026sciagenttrace} at commit \texttt{c36465d} and use its gpt-oss-120b LOO slice \cite{openai2025gptoss}: 93,282 actions, 6,663 problems, and 26,652 committee states across JEEBench \cite{arora2023jeebench}, LAB-Bench \cite{laurent2024labbench}, MaScQA \cite{zaki2024mascqa}, Omni-MATH-2 \cite{gao2025omnimath,ballon2026omnimath2}, and SciBench \cite{wang2023scibench}.  A \emph{role-only shell} has six empty fields but is rendered as a no-score notice plus role header; 1,584 are retained.  The headline matrix in \cite{yang2026wrongbutuseful} excludes them; the remaining 42-row difference reflects the v2.1.0 revision.  Effects comprise 5,431 harmful, 84,306 neutral, and 3,545 helpful rows for $K\in\{2,3,4,5\}$.  Source-problem hashes and exact duplicates remain in one benchmark-stratified 60/20/20 split; test has 1,332 problems and 5,328 committees.

Automated tests assert constant $y^{\rm all}$, $K$ LOO targets, signed-effect consistency, split exclusivity, provenance for 41 inputs, and absence of result fields.  Split files and executions are hashed; source-hash clustering gives CI $[+1.13,+2.50]$.

Before computing review-driven results, we hash-froze an input-validation-first protocol.  Shell committees are excluded from router fitting because remaining LOO actions are relative to the unfiltered prompt and cannot represent a second deletion.  We retrain and recalibrate on shell-free train/validation data.  At test, a validator removes one shell and stops; otherwise the router may remove one message.  The inspected test cache makes this review-driven, not a new untouched holdout.

For endpoint confirmation we froze all 158 validator actions plus 300 learned-tail actions (60 per benchmark).  Groq's public \texttt{openai/gpt-oss-120b} receives five calls per full/pruned prompt (temperature 0, low reasoning, 4,096-token cap): 4,580 integrations.  Calls were planned \texttt{full(r)} then \texttt{pruned(r)}; 450/458 committees preserve complete adjacent order, with retry/resume delays in eight.  Repeat IDs are not shared seeds, and this hosted backend is not the source endpoint.  The unweighted estimate describes only this sample.  For the source workload, validator actions receive weight one, learned-tail actions in benchmark $b$ receive $N_b/60$, and non-actions zero.  Before inspecting control outcomes, we froze a token-matched deletion control for the learned tail.

\subsection{Baselines and Metrics}
Offline baselines are Keep-All, random prune-one, lowest self-confidence/completeness, longest, and minority-answer pruning, thresholded to validation coverage.  Proposal-correctness and oracle LOO are diagnostics.  Metrics are accuracy, recoveries/corruptions, coverage, and official \texttt{HARMONY\_GPT\_OSS} tokens \cite{kundel2025gptoss}.  Endpoint estimates distinguish sample from design-weighted populations; accuracy is pairing-invariant, while corruption uses repeat-ID, independence-expected, and feasible-coupling definitions.  Problem-cluster bootstraps preserve all $K$.

\noindent\textbf{Procedure 1 (proposed target-backend qualification).}  On label-blind target calls, prune only if all of the following hold: (i) the design-weighted full-workload accuracy lower 95\% bound is at least $-1$ pp; (ii) acted-input reduction is at least 10\%; (iii) the weighted selector-minus-matched-control lower bound is nonnegative; (iv) the weighted expected-corruption UCL is at most 5\%; and (v) every prespecified benchmark, $K$, and completeness marginal on which the deployed pipeline acts has at least 59 acted problem clusters and UCL at most 5\%.  A zero-corruption group uses the larger of its clustered UCL and $1-.05^{1/n}$ over independent problem clusters.  Marginals with zero pipeline actions by design are exempt because \textsc{Keep-All} incurs no pruning risk.  The learned gate abstains at $K=2$, but the validator acts on 30 endpoint $K=2$ committees, so that marginal is not exempt here.  This is a global intersection--union decision \cite{berger1982}: any applicable failure or insufficient group returns \textsc{Keep-All}.  The complete rule was formulated after observing endpoint results and is applied here only retrospectively.

\section{Results}
\subsection{Overall Routing Performance}
Table~\ref{tab:main} reports sealed results.  SafeRoute gains 1.80 points, removes 12.76\% of tokens (12.75--12.77\% across serializations), and exceeds the random 100-seed expectation by 0.93 points (95\% CI $[0.46,1.42]$).  After Holm adjustment \cite{holm1979}, only differences versus random and minority-answer remain significant.

Lowest-self-confidence is statistically indistinguishable from SafeRoute ($+0.06$ points, CI $[-0.53,+0.66]$), but its 6.14\% helpful-prune UCL exceeds 5\%.  Completeness and longest pass; SafeRoute is one of several qualifying policies, with the highest point estimate, not uniquely safe.

The cached oracle reaches $+9.40$ points at 9.4\% coverage, quantifying recoverable headroom within the available single-deletion action space.

\begin{table}[t]
\centering
\caption{Previously sealed test routing results. Tok. is official specialist-message payload reduction; UCL is the 95\% problem-cluster bootstrap upper limit for helpful prunes/actions.}
\label{tab:main}
\resizebox{\columnwidth}{!}{%
\begin{tabular}{lrrrrrr}
\toprule
Method & Acc. & $\Delta$ pp & Rec./Corr. & Cov. & Tok. & UCL \\
\midrule
Keep-All & 74.89 & 0.00 & 0/0 & 0.0 & 0.0 & -- \\
SafeRoute-MAS & \textbf{76.69} & \textbf{+1.80} & 164/68 & 43.5 & 12.76 & \textbf{3.61} \\
Lowest self-conf. & 76.63 & +1.75 & 205/112 & 40.2 & 12.07 & 6.14 \\
Completeness & 76.13 & +1.24 & 142/76 & 39.8 & 11.08 & 4.34 \\
Longest & 76.11 & +1.22 & 145/80 & 40.1 & \textbf{13.59} & 4.50 \\
Minority answer & 75.51 & +0.62 & 164/131 & 39.4 & 12.75 & 7.28 \\
Random (100 seeds)$^a$ & 75.76 & +0.87 & 121.9/75.6 & 40.8 & 11.68 & -- \\
\midrule
Proposal-correct. (diag.) & 77.05 & +2.16 & 205/90 & 40.6 & 12.33 & 4.94 \\
Oracle LOO (cached upper) & 84.29 & +9.40 & 501/0 & 9.4 & 2.53 & 0.00 \\
\bottomrule
\end{tabular}}
\vspace{0.5mm}
\parbox{\columnwidth}{\footnotesize $^a$The mean random policy has no single selected-action set, so no UCL is assigned.  Rows below the rule are non-deployable diagnostics using gold correctness or realized LOO labels.}
\end{table}

\subsection{Input Validation, Missingness, and Token Accounting}
\label{sec:missingness}
In the original frozen policy, 53 of 96 net recoveries arise in 177 schema-incomplete committees, and LAB-Bench supplies 88; 4,673 complete non-LAB committees have 44 recoveries and 44 corruptions.  Across the action table, shell labels are asymmetric (673 harmful versus 25 helpful), but the independent-endpoint validator branch changes accuracy by $-0.25$ points, so this effect does not reproduce.

The frozen review experiment uses a one-shell validator (shortest shell; index tie-break) and a shell-free learned model.  The tail retains $+0.87$ points on 5,170 committees; the exact cascade gains $+1.91$, adding $+0.84$ beyond validation (95\% CI $[+.19,+1.52]$).  The validator's 8 corruptions among 158 actions yield 5.06\% conditional risk and 8.78\% UCL.  Thus cleanup helps in-cache but fails the same conditional-risk criterion.

Conditional risk counts corruptions per action; committee risk counts harmed committees.  Table~\ref{tab:safety} reports both together with coverage.  The aggregate frozen policy does not certify every subgroup: LAB-Bench and incomplete committees exceed the 5\% UCL, as do 8 acted-on cells in the complete 36-cell benchmark$\times K\times$completeness analysis.  The ordinary bootstrap assigns zero to zero-event cells; our retrospective reanalysis instead uses the exact problem-cluster safeguard above.  Of 36 cross-cells, 9 have no action, 17 have insufficient support, 6 pass, and 4 fail risk.  A validation-only guard improves the cached aggregate but does not eliminate LAB's violation; subgroup gating is not a distribution-free guarantee.

\begin{table}[t]
\centering
\caption{Frozen-primary subgroup risk and coverage. Risk and UCL are conditional helpful prunes/actions; Corr. is corruptions/all. LOBO $\Delta$ uses the shell-free, no-benchmark-feature held-out experiment. Percentages throughout.}
\label{tab:safety}
\resizebox{\columnwidth}{!}{%
\begin{tabular}{lrrrrr}
\toprule
Scope & Cov. & Risk & UCL & Corr. & LOBO $\Delta$ pp \\
\midrule
JEEBench & 50.0 & 0.49 & 1.46 & 0.24 & +0.05 \\
LAB-Bench & 74.2 & 4.78 & 6.79 & 3.55 & +8.76 \\
MaScQA & 36.2 & 2.13 & 4.44 & 0.77 & -0.04 \\
Omni-MATH-2 & 41.2 & 2.97 & 3.88 & 1.23 & -0.45 \\
SciBench & 22.8 & 0.95 & 2.75 & 0.22 & -0.09 \\
\midrule
$K=2$ & 0.0 & -- & -- & 0.00 & -- \\
$K=3$ & 49.5 & 1.52 & 2.35 & 0.75 & -- \\
$K=4$ & 59.5 & 3.28 & 4.36 & 1.95 & -- \\
$K=5$ & 64.9 & 3.70 & 4.77 & 2.40 & -- \\
Complete & 42.2 & 2.90 & 3.61 & 1.22 & -- \\
Incomplete & 79.7 & 3.55 & 6.21 & 2.82 & -- \\
\bottomrule
\end{tabular}}
\end{table}

At the frozen policy's validation token budget, realized cached-test reductions are 11.91--12.76\% across policies; savings are therefore not unique to SafeRoute.  Confidence attains $+2.03$ points but its 5.27\% helpful-prune UCL exceeds the criterion.  SafeRoute attains $+1.80$ points (UCL 3.61\%), compared with completeness at $+1.50$ (4.30\%) and longest at $+1.24$ (4.07\%).

\subsection{Ablation and Sensitivity}
\label{sec:ablation}
Matched-coverage ablations are post-hoc.  Metadata alone matches the primary model.  Removing masks gives $+1.60$ points; removing context, benchmark, calibration, or diversity gives $+2.08$, $+1.63$, $+1.41$, or $+2.44$.  No component is causally necessary; context and diversity remain frozen engineering choices.

Variant UCLs are 2.99--3.91\%.  On acted candidates, predicted versus observed helpful risk is 2.54\% versus 2.93\% (ECE 0.61\%).

\subsection{Benchmark and Committee-Size Robustness}
\begin{figure}[t]
    \centering
    \includegraphics[width=0.90\columnwidth]{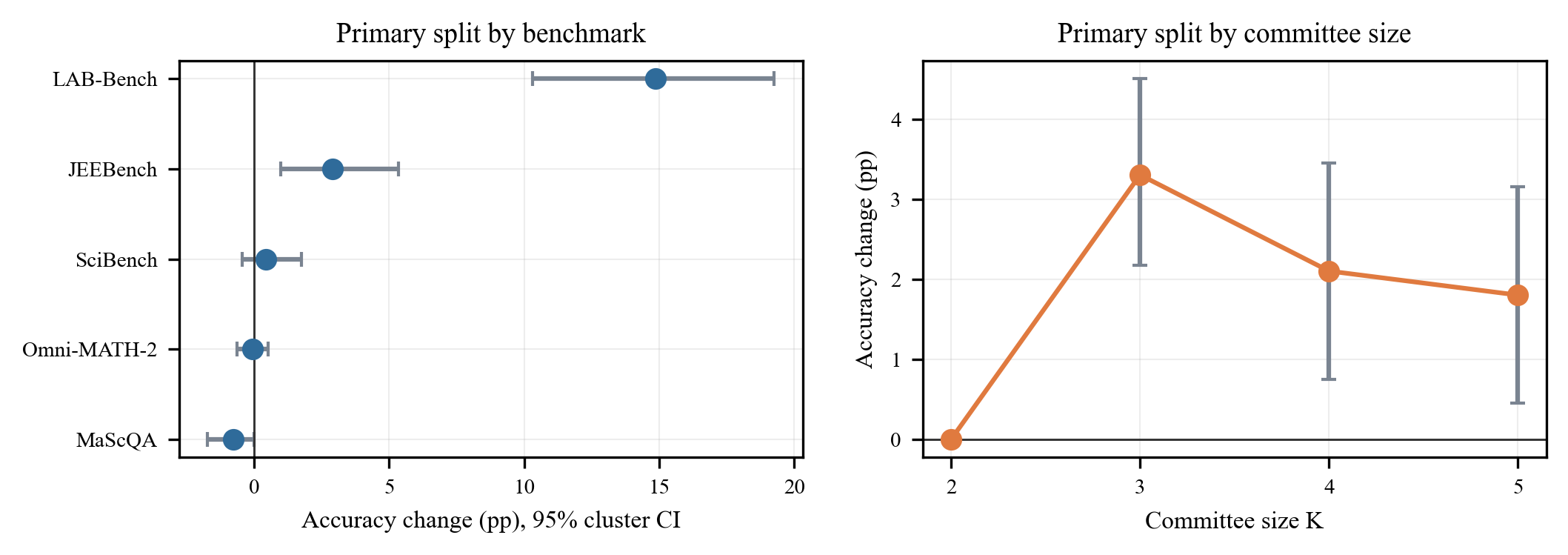}
    \caption{Cached primary-split accuracy changes by benchmark and committee size (problem-cluster 95\% CIs).  Heterogeneity motivates target-specific qualification; $K=2$ is a frozen no-action boundary.}
    \label{fig:robustness}
\end{figure}

Figure~\ref{fig:robustness} and Table~\ref{tab:safety} show strong heterogeneity: LAB-Bench supplies 88 of 96 original net recoveries, whereas the non-LAB gain is $+0.17$ (95\% CI $[-0.30,+0.63]$).  Shell-free LOBO is clearly positive only for LAB.  LAB has the lowest base accuracy and highest shell-committee rate, but is neither the longest nor most answer-divergent benchmark; no observed property defines a universal portability rule.

Across $K=3,4,5$, the primary gains are 3.30, 2.10, and 1.80 points.  A model trained on $K\le4$ acts on every $K=5$ committee, so its threshold does not transfer across $K$.

In 478 complete LAB committees, SafeRoute acts 349 times (62 recoveries, 19 corruptions), a 5.4\% conditional risk.  Post-hoc role/length diagnostics do not reproduce its 43 net recoveries: the strongest fixed-role rule yields only $+2.93$ points.  Thus the aggregate criterion does not certify this stratum, and its mechanism remains unresolved.

\subsection{Retrospective Endpoint Qualification Diagnostic}
Table~\ref{tab:confirmation} separates sample and target-population estimands.  The frozen sample gives $-0.087$ points, but it censors non-actions, includes every validator action, and balances the learned tail across benchmarks.  Design weighting over 158 validator and 3,599 learned-tail actions, with zero effect assigned to non-actions, gives $+0.943$ points on the 5,328-committee workload (95\% CI $[-1.315,+3.501]$) and 18.72\% acted-input reduction.  Thus neither benefit nor noninferiority is established.  The weighted tail-minus-matched estimate is $+1.680$ points (95\% CI $[-1.270,+5.098]$); weighted token reductions are 19.31\% and 17.68\%, respectively, so selector advantage is also unestablished.

\begin{table}[t]
\centering
\caption{Retrospective endpoint diagnostic. Wtd. rows use source-workload design weights; Tok. is acted integrator-input reduction. CIs/UCLs use problem clusters.}
\label{tab:confirmation}
\resizebox{\columnwidth}{!}{%
\begin{tabular}{lrrrr}
\toprule
Estimand & Units & $\Delta$ pp (95\% CI) & Tok. & Risk UCL \\
\midrule
Frozen 458 sample & 2,290 & $-0.09$ $[-1.85,1.61]$ & 12.63 & 5.37$^a$ \\
Wtd. full workload & 5,328 & $+0.94$ $[-1.31,3.50]$ & 18.72 & 4.60$^b$ \\
Validator branch & 790 & $-0.25$ $[-2.84,2.14]$ & 3.97 & 7.23$^b$ \\
Balanced learned tail & 1,500 & $0.00$ $[-2.32,2.30]$ & 16.81 & 4.94$^b$ \\
Wtd. tail vs matched & 3,599 & $+1.68$ $[-1.27,5.10]$ & +1.63 & -- \\
\bottomrule
\end{tabular}}
\vspace{0.4mm}
\parbox{\columnwidth}{\footnotesize $^a$Repeat-ID pairing; $^b$independence-expected corruption. Tok. for the full-workload row is among routed actions.}
\end{table}

The accuracy difference is pairing invariant, but corruption counts are not.  Repeat-ID pairing gives 4.15\% risk and 5.37\% UCL; five cyclic pairings give UCLs 5.02--5.36\%, independence-expected corruption gives 3.98\% and 5.24\%, and feasible coupling UCLs span 4.34--6.06\%.  After design weighting, expected risk is 3.05\% (UCL 4.60\%).  By five-call majority, only 13 of 88 cached non-neutral labels (14.8\%) retain the same direction; backend shift and repeated-call variability cannot be separated.  Of 11 applicable endpoint marginals, 2 pass, five lack 59 problem clusters, and four additional, sufficiently supported marginals exceed the 5\% UCL.  Procedure~1 therefore rejects Groq on accuracy and matched-control; the subgroup gate also fails from both insufficient evidence and observed excess risk.  The verdict is unchanged for accuracy margins 0.5--1.5 pp, risk limits 4--6\%, and token floors 5--15\%.

\section{Discussion and Limitations}
SafeRoute combines a deployment-visible representation with a calibrated selective gate.  Each decision uses its own cached replay; no result sums single-message effects.  Metadata-only performance and the context/diversity ablations make this a scoped routing method, not evidence that every advertised component independently helps.

The validator-first rerun separates cleanup from routing within the single-deletion action space.  It supports a positive tail in the inspected cache, but endpoint repetition does not confirm it; unvalidated groups should abstain.

The repeated experiment uses a new hosted backend; it cannot isolate endpoint shift from sampling variation, and temperature-zero calls are not seed-controlled.  The 1-pp accuracy margin, 10\% token floor, and 5\% risk tolerance are author-declared engineering values.  We distinguish demonstrated diagnostics from deployment claims: Procedure~1 was formalized after observing the endpoint experiment, its Groq application is retrospective, and prospective validation remains future work.  As a proposed future deployment rule, SafeRoute defaults to Keep-All unless the exact target passes every gate globally.

\section{Conclusion}
SafeRoute identifies a positive source-compatible cached operating point, but neither the balanced endpoint sample nor its design-weighted workload estimate confirms an accuracy gain or selector advantage.  Its applied contribution is a numerical target-specific qualification and abstention procedure, demonstrated here retrospectively: validate inputs, calibrate on the deployment backend, compare with matched pruning, check aggregate and subgroup risks, and otherwise retain Keep-All.

\begingroup
\footnotesize
\bibliographystyle{IEEEtran}
\bibliography{references}
\endgroup
\end{document}